\documentclass[article,10pt]{IEEEtran}
\usepackage[named]{algo}
\usepackage{graphics}
\usepackage{graphicx}
\usepackage{stfloats}
\usepackage{amssymb}
\usepackage{amsmath}
\usepackage{amsfonts}
\usepackage{subfigure}
\usepackage{float}
\usepackage{pifont}
\usepackage{setspace}
\usepackage{url}

\newcommand{\Section}{\section}
\newcommand{\SubSection}{\subsection}

\def\urltilda{\kern -.15em\lower .7ex\hbox{\~{}}\kern .04em}
\def\urldot{\kern -.10em.\kern -.10em}
\def\urlhttp{http\kern -.10em\lower -.1ex\hbox{:}\kern -.12em\lower 0ex\hbox{/}\kern -.18em\lower 0ex\hbox{/}}

\begin{document}

\title{Binary Fused Compressive Sensing: 1-Bit Compressive Sensing meets Group Sparsity}

\author{Xiangrong Zeng\ \  and \  M\'{a}rio A. T. Figueiredo  \thanks{Manuscript submitted December 20, 2012. } \\ \thanks{ Both authors are with the Instituto de Telecomunica\c{c}\~oes and the Department of Electrical and Computer Engineering, Instituto Superior T\'ecnico, Technical University of Lisbon, 1049-001, Lisboa, Portugal.
Email: Xiangrong.Zeng@lx.it.pt, mario.figueiredo@lx.it.pt. }}

\maketitle

\begin{abstract}
We propose a new method, {\it binary fused compressive sensing} (BFCS), to recover sparse piece-wise smooth signals from 1-bit compressive measurements.
The proposed algorithm is a modification of the previous {\it binary iterative hard thresholding} (BIHT) algorithm, where, in addition to the sparsity constraint, the total-variation of the recovered signal is upper constrained. As in BIHT, the data term of the objective function is an one-sided $\ell_1$ (or $\ell_2$) norm.
Experiments on the recovery of sparse piece-wise smooth signals show that the proposed algorithm is able to take advantage of the piece-wise smoothness of the original signal, achieving more accurate recovery than BIHT.
\end{abstract}


\Section{Introduction}\label{sec:intro}

In classic {\it compressive sensing} (CS) \cite{candes2006stable}, \cite{donoho2006}, a sparse signal ${\bf x} \in {\mathbb R}^n$ is shown to be recoverable from a few  linear measurements
\begin{equation}\label{linearmodel}
 {\bf b}={\bf A}{\bf x},
\end{equation}
where  ${\bf b}\in\mathbb{R}^{m}$ is the measurement vector, ${\bf A}\in\mathbb{R}^{m\times n}$ is a known sensing matrix (that must satisfy some conditions), and the fact that $ m \ll n$ leads to the ill-posed nature of (\ref{linearmodel}). This formulation assumes that the measurements are real-valued, thus ignoring that, in practice, any acquisition involves quantization; when quantization is taken into account, we have {\it quantized CS} (QCS) \cite{sun2009quantization}, \cite{zymnis2010compressed}, \cite{laska2011democracy}, \cite{jacques2011dequantizing}, \cite{laska2012regime}, \cite{liu2012robust}. The interesting extreme case of QCS is  1-bit CS \cite{boufounos20081},
\begin{equation}\label{1bitcs}
 {\bf y}=\mbox{sign}\left({\bf A}{\bf x}\right),
\end{equation}
where $\mbox{sign}(\cdot)$ is the element-wise sign function, which returns $+1$ for positive arguments and $-1$ otherwise. Such 1-bit measurements can be acquired by a comparator with zero, which is very inexpensive and fast, as well as robust to amplification distortion, as long as the signs of the measurements are preserved. In contrast with the measurement model of conventional CS, 1-bit measurements only keep the signs, thus loosing any information about the magnitude of the original signal ${\bf x}$. The goal is then to recover ${\bf x}$, but only up to an unknown and unrecoverable magnitude.

The first algorithm for recovering a sparse or compressible signal from 1-bit measurements (named {\it renormalized fixed point iteration} -- RFPI)
was proposed by Boufounos and Baraniuk \cite{boufounos20081}. Boufounos later showed that recovery from nonlinearly distorted measurements is also possible, even if the nonlinearity is unkonwn \cite{boufounos2010reconstruction}, and introduced a greedy algorithm ({\it matching sign pursuit} -- MSP) \cite{boufounos2009greedy}. After that seminal work, several algorithms for 1-bit CS have been developed; a non-comprehensive list includes linear  programming  \cite{plan2011one}, \cite{plan2012robust}, the {\it restricted-step shrinkage} (RSS) \cite{laska2011trust}, and {\it binary iterative hard thresholding} (BIHT) \cite{jacques2011robust}. BIHT has a simple form and performs better than the previous algorithms, both in terms of reconstruction error and consistency. More recently, \cite{fang2012fast} and \cite{kamilov2012one} proposed new algorithms for 1-bit CS, based on {\it generalized approximate message passing} (GAMP) \cite{rangan2011generalized} and {\it majorization-minimization} (MM) \cite{hunter2004tutorial}, \cite{figueiredo2005bound} methods, respectively. Considering the sign flips caused by noise, \cite{yan2012robust} introduced a technique called {\it adaptive outlier pursuit} (AOP), and \cite{movahed2012robust} extended it by proposing an algorithm termed {\it noise-adaptive RFPI } (NARFPI). Finally, \cite{bourquard2012binary} and \cite{yang2012bits} applied 1-bit CS in image acquisition and \cite{davenport20121} studied matrix completion from noisy 1-bit observations.

In this paper, we will focus on the topic of recovering group-sparse signals from 1-bit CS measurements. The rationale is that group-sparsity expresses more structured knowledge about the unknown signal than simple sparsity, thus potentially allowing for more robust recovery from fewer measurements. To our knowledge, there is no previous literature on this topic.  In recent years, several group-sparsity-inducing regularizers have been proposed, including {\it group LASSO} (gLASSO) \cite{yuan2005model}, {\it fused LASSO} (fLASSO) \cite{tibshirani2004sparsity},  {\it elastic net} (EN) \cite{zou2005regularization}, {\it octagonal shrinkage and clustering algorithm for regression} (OSCAR) \cite{bondell2007simultaneous}, {\it sparse group LASSO} (sgLASSO) \cite{simon2012sparse}, {\it weighted fLASSO} (wfLASSO) \cite{daye2009shrinkage}, {\it graph-guided fLASSO} (ggfLASSO) \cite{kim2009multivariate}, and {\it adaptive elastic net } (aEN) \cite{zou2009adaptive}. Wheras the gLASSO, sgLASSO and ggfLASSO need prior knowledge of the structure of groups, which is a strong requirement in many applications, the EN, aEN, and OSCAR regularizers (which do not require this prior knowledge) group the variables based only the signal magnitudes, which is naturally unsuitable for 1-bit CS. The fLASSO and wfLASSO promote grouping through penalizing the difference between each variable and its neighbors,  and their penalty terms are in fact similar to a one-dimensional {\it total variation} (TV) norm \cite{ROF1992}.

From the discussion in the previous paragraph, it can be seen that TV regularization has the potential to be applied in recovering sparse and piece-wise smooth signals from 1-bit CS observations. In this paper, we combine the advantages of the BIHT algorithm and TV regularization, leading to what we call the {\it binary fused compressive sensing} (BFCS) regularizer, which promotes grouping in 1-bit CS reconstruction. Observe that the problem of detecting sign flips (errors) is not considered in this paper, but it will be addressed in a longer upcoming version.

The rest of the paper is organized as follows. Section II describes the BFCS formulation and algorithm, Section III reports experimental
results, and Section IV concludes the paper.

\Section{Binary Fused Compressive Sensing (BFCS)}
\SubSection{The Observation Model}
In this paper, we consider the noisy 1-bit measurement model,
\begin{equation}\label{1bitcsnoisy}
 {\bf y}=\mbox{sign}\left({\bf A}{\bf x} + \boldsymbol {\bf w}\right),
\end{equation}
where ${\bf y}\in \left\{+1, -1\right\}^{m}$,  ${\bf A}\in\mathbb{R}^{m\times n}$ is the sensing matrix, ${\bf x} \in {\mathbb R}^n$ is the original sparse piece-wise signal, and $\boldsymbol {\bf w} \in {\mathbb R}^m $ represents additive white Gaussian noise with the variance  $\sigma^2$.
The following subsections review the  BIHT algorithm and introduce our proposed BFCS method for recovering ${\bf x}$ from ${\bf y}$.
\SubSection{Binary Iterative Hard Thresholding (BIHT)}
To recover ${\bf x}$ from ${\bf y}$, Jacques {\it et al} \cite{jacques2011robust} proposed the criterion
\begin{equation}
\label{BIHT}
\begin{split}
& \min_{\bf x}  f({\bf y}\odot {\bf A}{\bf x}) + \iota_{\Sigma_K}({\bf x})\\
& \mbox{subject\; to }\; \left\|{\bf x} \right\|_2 = 1,
\end{split}
\end{equation}
where:  the operation ``$\odot$" represents the element-wise product;
$\iota_C\left({\bf x}\right)$ denotes the indicator function of set $C$,
\begin{equation}
\iota_C\left({\bf x}\right) = \left\{
\begin{array} {ll}
0,	&  {\bf x} \in C \\
+\infty, & {\bf x} \not\in C;
\end{array}\right.
\label{indicator}
\end{equation}
${\Sigma_K} = \left\{{\bf x}\in \mathbb{R}^n : \left\|{\bf x}\right\|_0 \leq K \right\}$ (with
$\|{\bf v}\|_0$ denoting the number of non-zero components in ${\bf v}$) 
is the set of $K$-sparse signals; and $f$ is one of the penalty functions
defined next. To penalize linearly the violations of the sign consistency between the observations and the estimate, the choice is
$f({\bf z}) = 2\left\|{\bf z}_-\right\|_1$, where ${\bf z}_- = \min \left({\bf z}, 0\right)$ (where the minimum is
applied component-wise and the factor 2 is included for later convenience) and $\|{\bf v}\|_1 = \sum_i |v_i|$ is the
$\ell_1$ norm.
 Quadratic penalization of the sign violations
is achieved by using  $f({\bf z})=\frac{1}{2} \left\|{\bf z}_-\right\|_2^2 $, where the factor $1/2$ is also
included for convenience. The iterative hard thresholding (IHT) \cite{IHT2009} algorithm applied to \eqref{BIHT}
(ignoring the norm constraint during the iterations) leads to the BIHT algorithm \cite{jacques2011robust}:
\vspace{0.1cm}
\begin{algorithm}{BIHT}{
\label{alg:BIHT}}
set $k =0, \tau >0, {\bf x}_0$ and $K$ \\
\qrepeat\\
     ${\bf v}_{k+1} = {\bf x}_{k} - \tau \partial f\left({\bf y}\odot {\bf A}{\bf x}_k\right)$\\
		 ${\bf x}_{k+1} =  {\mathcal P}_{\Sigma_K} \left( {\bf v}_{k+1}\right)$\\
		 $ k \leftarrow k+ 1$
\quntil some stopping criterion is satisfied.\\
\qreturn ${\bf x}_k/\left\|{\bf x}_k\right\|$
\end{algorithm}
\vspace{0.2cm}
In this algorithm, $\partial f$
denotes the subgradient of the objective (see \cite{jacques2011robust}, for details),
which is given by
\begin{equation}
\partial f\left({\bf y}\odot {\bf A}{\bf x}\right) = \left\{
\begin{array} {ll}
{\bf A}^T\left(\mbox{sign}({\bf A}{\bf x}) - {\bf y}\right),	& \mbox{$\ell_1$ objective} \\
\left({\bf Y}{\bf A}\right)^T\left({\bf Y}{\bf A}{\bf x}\right)_-, & \mbox{$\ell_2$ objective} ,
\end{array}\right.
\label{subgradient}
\end{equation}
where ${\bf Y} = \mbox{diag}({\bf y})$ is a diagonal matrix with vector ${\bf y}$ in its diagonal.
Step 3 corresponds to a sub-gradient descent step (with step-size $\tau$), while
Step 4 performs the projection onto the non-convex set ${\Sigma_K}$, which 
corresponds to computing the best $K$-term approximation of ${\bf v}$, that is, 
keeping $K$ largest components in magnitude and setting the others to zero. Finally, the 
returned solution is projected onto the unit sphere to satisfy the constraint $\left\|{\bf x} \right\|_2 = 1$ in (\ref{BIHT}).
The versions of BIHT with $\ell_1$ and $\ell_2$ objectives are referred to as BIHT and BIHT-$\ell_2$, respectively.

\SubSection{Proposed formulation and algorithm}
The proposed formulation essentially adds a new constraint of
low total variation to the criterion \eqref{BIHT},
\begin{equation}\label{BFCS}
\begin{split}
& \min_{\bf x}  f\left({\bf y}\odot {\bf A}{\bf x}\right) + \iota_{\Sigma_K} \left({\bf x}\right) + \iota_{F_{\epsilon}} \left({\bf x}\right)\\
& \mbox{subject to} \; \left\|{\bf x} \right\|_2 = 1,
\end{split}
\end{equation}
where $F_{\epsilon}= \left\{{\bf x}\in \mathbb{R}^n :\; \mbox{TV} \left({\bf x}\right)  \leq \epsilon \right\}$, with 
$\mbox{TV} \left({\bf x}\right) $ denoting the total variation (TV), which in the one-dimensional case is defined as
\begin{equation}
\mbox{TV}({\bf x}) = \sum_{i=1}^{n-1} |x_{i+1} - x_i|,
\end{equation}
and $\epsilon$ is a positive parameter. Notice that the vectors in $F_{\epsilon}$ are characterized by having consecutive variables with not too dissimilar values, which justifies the term ``fused". The proposed regularizer ${\phi}_{BFCS}\left({\bf x}\right) = 
\iota_{\Sigma_K} \left({\bf x}\right) + \iota_{F_{\epsilon}} \left({\bf x}\right)$ simultaneously promotes sparsity and grouping.

In the same vein as BIHT, the proposed algorithm is as follows:
\vspace{0.3cm}
\begin{algorithm}{BFCS}{
\label{alg:BFCS}}
set $k =0, \tau >0, \epsilon>0, {\bf x}_0$ and $K$ \\
\qrepeat\\
     ${\bf v}_{k+1} = {\bf x}_{k} - \tau \partial f\left({\bf y}\odot {\bf A}{\bf x}_k\right)$\\
		  ${\bf x}_{k+1} =  {\mathcal P}_{\Sigma_K} \! \left(   {\mathcal P}_{F_{\epsilon}} ( {\bf v}_{k+1})\right)$\\
		  $ k \leftarrow k+ 1$
\quntil some stopping criterion is satisfied.\\
\qreturn ${\bf x}_k/\left\|{\bf x}_k\right\|$
\end{algorithm}
\vspace{0.3cm}
The projection onto $F_{\epsilon}$ is computed using the algorithm proposed by Fadili and Peyr\'e \cite{fadili2011total}. 
Of course, the objective function in \eqref{BFCS} is not convex (since $\Sigma_K$ is not a convex set and the $\{{\bf x}\in \mathbb{R}^n:
\; \|{\bf x}\|_2=1\}$ is also not a convex set), thus there is no guarantee that the algorithm finds a global 
minimum. The versions of the
BFCS algorithm with $\ell_1$ and $\ell_2$ objectives are referred to as BFCS and BFCS-$\ell_2$, respectively.

If the original signal is known  to be non-negative, then the algorithm should include  a projection onto $\mathbb{R}_+^n$ in each iteration.

\Section{Experiments}
In this section, we report results of experiments aimed at comparing the performance of BFCS with that of BIHT.
All the experiments were performed using MATLAB on a 64-bit Windows 7 personal computer with an Intel Core i7 3.07 GHz processor and 6.0 GB of RAM.
In order to measure the performance of different algorithms, we employ the following five metrics defined on an estimate ${\bf e}$ of an original vector ${\bf x}$.

\begin{itemize}
	\item Mean absolute error, $\textbf{MAE} = \left\|{\bf x} - {\bf e}\right\|_1/n$; 
	\item Mean square error, $\textbf{MSE} = \left\|{\bf x} - {\bf e}\right\|^2/n$;
	\item Signal-to-noise ratio (notice that $\|{\bf x}\|_2^2=1$), $\textbf{SNR} = -10 \log_{10} \left( \left\|{\bf x} - {\bf e}\right\|^2_2 \right) $;
  \item Position error rate, $\textbf{PER}  = \sum_i \bigl\lvert |\mbox{sign}(x_i)| - |\mbox{sign}( e_i )  |  \bigr\rvert /n$;	
	\item Angle error (notice that $\|{\bf x}\|_2^2 =\|{\bf e}\|_2^2 =1$), $\textbf{AGE}  = \arccos\left\langle {\bf x}, {\bf e}\right\rangle / \pi$.
\end{itemize}

The original signals ${\bf x}$ are taken as sparse and piece-wise smooth, of length $n = 2000$, with two alternative sparsity levels 
$K = 100$ or $400$; specifically, $\bar{\bf x}$ is generated as
 \begin{equation} \label{piecewisesignal}
\bar{x}_i =\left\{ \begin{array}{lll}
2 + 0.05 \, k_i, & i \in \mathcal{B}\\
-1 + 0.05 \, k_i, & i \in \mathcal{C} \\
0, & i \not \in \left(\mathcal{B} \cup \mathcal{C} \right)
\end{array} \right.
\end{equation}
where the $k_i$ are independent samples of a zero-mean, unit variance Gaussian random variable,  
and the sets $\mathcal{B}$ and $\mathcal{C}$ are defined as
\begin{align}
\mathcal{B} & = \left\{ 100, \cdots, 100 + K/4 - 1  , 500, \cdots, 500 + K/4 -1 \right\}\nonumber\\
\mathcal{C} & = \left\{ 1000, \cdots, 1000 + K/4 -1  , 1500, \cdots, 1500 + K/4 -1  \right\}\nonumber ;
\end{align}
the signal is then normalized, ${\bf x} = \bar{\bf x}/\|\bar{\bf x}\|$. The sensing matrix ${\bf A}$ is a $1000 \times 2000$ matrix with components sampled from the standard normal distribution. Finally, observations ${\bf y}$ are obtained by \eqref{1bitcsnoisy}, with noise standard deviation $\sigma = 1$ or $4$.

\begin{figure*}
	\centering
		\includegraphics{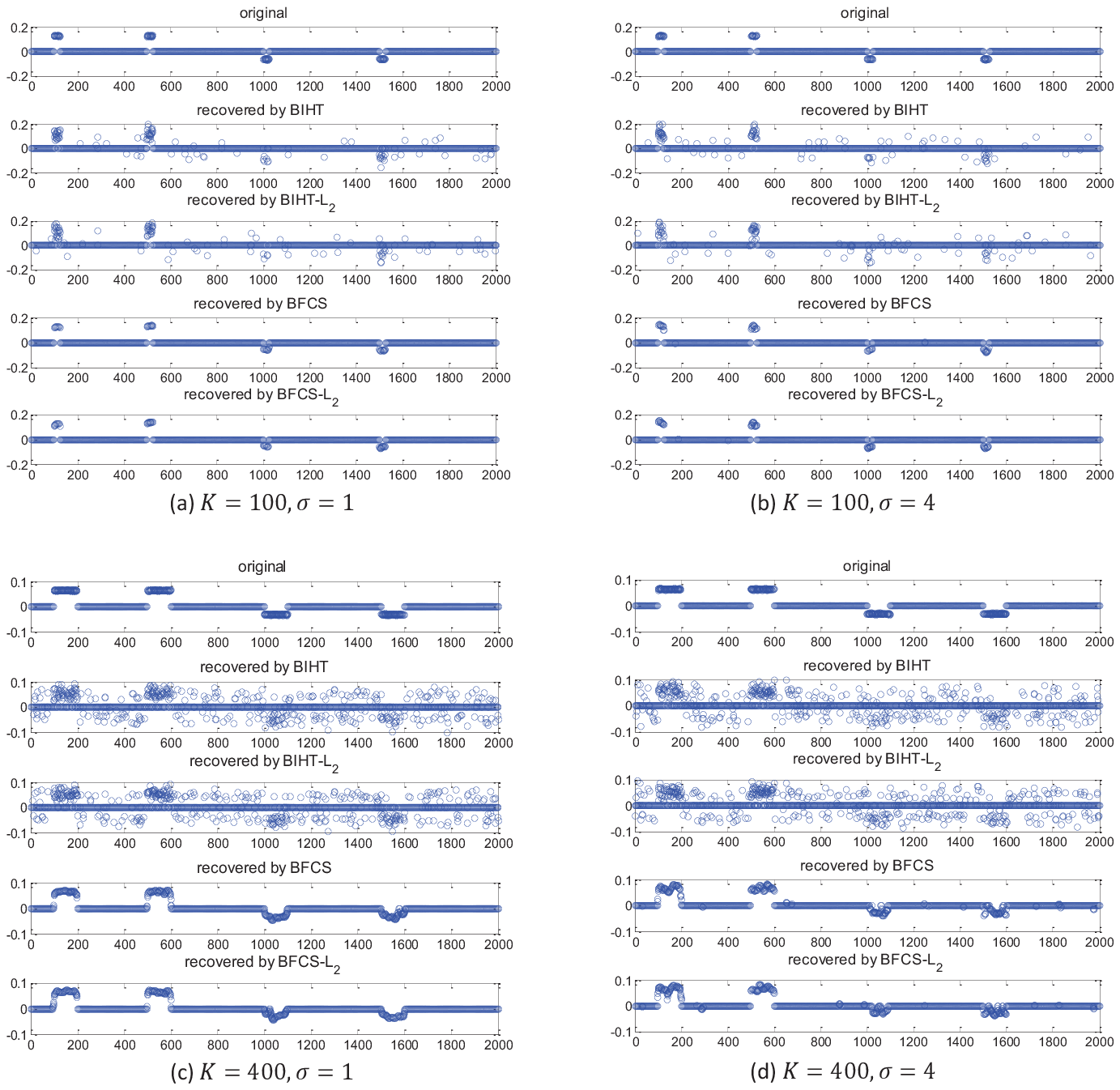}
	\caption{Recovered sparse piece-wise signals by different algorithms over $K$ and $\sigma$}
	\label{fig:differentK}
\end{figure*}

We run the algorithms BIHT, BIHT-$\ell_2$, BFCS and BFCS-$\ell_2$ for all four combinations of sparsity and noise level: $K = 100,\; \sigma = 1$; $K =100,\; \sigma = 4$; $K = 400,\; \sigma = 1$; $K = 400,\; \sigma =4$. The stopping criterion is  $\left\|{\bf x}_{(k+1)}- {\bf x}_{(k)}\right\|/\left\|{\bf x}_{(k+1)}\right\|\leq 0.001$, where ${\bf x}_{(k)}$ is estimate at the $k$-th iteration. In BIHT and BIHT-$\ell_2$, the step-size is $\tau = 1$ (setup of \cite{jacques2011robust}), while in BFCS and BFCS-$\ell_2$, $\tau$ and $\epsilon$ are hand tuned in each case for the best improvement in SNR. The recovered signals are shown in Figure \ref{fig:differentK} and the results of MAE, MSE, SNR, PER and AGE are shown in Table \ref{tab:resultsofmetrics}.

From Table \ref{tab:resultsofmetrics} and Figure \ref{fig:differentK}, we can  see that BFCS and BFCS-$\ell_2$ perform better than BIHT and BIHT-$\ell_2$, in
terms of MAE, MSE, SNR, PER and AGE. This advantage is larger for larger values of $K$ and $\sigma$. Moreover, BFCS performs better than BFCS-$\ell_2$ at low noise level ($\sigma = 1$), while BFCS-$\ell_2$ outperforms BFCS at high noise level ($\sigma = 4$).

\Section{Conclusions}
We have proposed an algorithm for recovering  sparse piece-wise smooth signals from 1-bit compressive measurements. We have shown that if the original signals are in fact sparse and piece-wise smooth, the proposed method (termed BFCS -- {\it binary fused hard thresholding}) outperforms (under several accuracy measures) the previous method BIHT ({\it binary iterative hard thresholding}), which  relies only on sparsity of the original signal.  Future work will involve using the technique of detecting sign flips to obtain a robust version of BFCS.

\begin{table}
\centering \caption{Results of metrics} \label{tab:resultsofmetrics}
\begin{tabular}{|p{0.8cm}|l|l|l|l|p{1.05cm}|}
\hline
\footnotesize Metrics &   \footnotesize $K$ and $\sigma $ & \footnotesize BIHT & \footnotesize BIHT-$\ell_2$ & \footnotesize BFCS & \footnotesize BFCS-$\ell_2$ \\ \hline \hline
                  & \footnotesize $K = 100$, $\sigma = 1$ & \footnotesize 3.23E-4 & \footnotesize 3.43E-3 & \footnotesize 2.74E-4 & \footnotesize 4.59E-4 \\ \cline{2-6}
\footnotesize MAE & \footnotesize $K = 100$, $\sigma = 4$ & \footnotesize 3.45E-3 & \footnotesize 3.72E-3 & \footnotesize 6.72E-4 & \footnotesize 6.38E-4 \\ \cline{2-6}
                  & \footnotesize $K = 400$, $\sigma = 1$ & \footnotesize 1.14E-2 & \footnotesize 1.15E-2 & \footnotesize 1.26E-3 & \footnotesize 1.68E-3  \\ \cline{2-6}
                  & \footnotesize $K = 400$, $\sigma = 4$ & \footnotesize 1.19E-2 & \footnotesize 1.17E-2 & \footnotesize 1.96E-3 & \footnotesize 1.87E-3 \\ \hline \hline
									
                  & \footnotesize $K = 100$, $\sigma = 1$ & \footnotesize 2.07E-4 & \footnotesize  2.25E-5& \footnotesize 2.50E-6 & \footnotesize  8.50E-6\\ \cline{2-6}
\footnotesize MSE & \footnotesize $K = 100$, $\sigma = 4$ & \footnotesize 2.36E-4 & \footnotesize  2.69E-5& \footnotesize 1.79E-5 & \footnotesize  1.73E-5\\ \cline{2-6}
                  & \footnotesize $K = 400$, $\sigma = 1$ & \footnotesize 5.12E-4 & \footnotesize  5.15E-4& \footnotesize 1.73E-5 & \footnotesize  2.54E-5\\ \cline{2-6}
                  & \footnotesize $K = 400$, $\sigma = 4$ & \footnotesize 5.33E-4 & \footnotesize  5.29E-4& \footnotesize 3.29E-5 & \footnotesize  3.16E-5\\ \hline \hline		
																
                  & \footnotesize $K = 100$, $\sigma = 1$ & \footnotesize 3.82 & \footnotesize 3.48 & \footnotesize 23.0 & \footnotesize 17.7 \\ \cline{2-6}
\footnotesize SNR & \footnotesize $K = 100$, $\sigma = 4$ & \footnotesize 3.25 & \footnotesize 2.69 & \footnotesize 14.1 & \footnotesize 14.5 \\ \cline{2-6}
                  & \footnotesize $K = 400$, $\sigma = 1$ & \footnotesize -0.102 & \footnotesize -0.126 & \footnotesize 14.6 & \footnotesize 12.9 \\ \cline{2-6}
                  & \footnotesize $K = 400$, $\sigma = 4$ & \footnotesize -0.274 & \footnotesize  -0.271& \footnotesize 11.8 & \footnotesize 12.0 \\ \hline \hline															 

                  & \footnotesize $K = 100$, $\sigma = 1$ & \footnotesize 3.7\% & \footnotesize 4.0\% & \footnotesize 0 & \footnotesize 0.2\% \\ \cline{2-6}
\footnotesize PER & \footnotesize $K = 100$, $\sigma = 4$ & \footnotesize 3.9\% & \footnotesize 4.2\% & \footnotesize 0.6\%& \footnotesize 0.4\% \\ \cline{2-6}
                  & \footnotesize $K = 400$, $\sigma = 1$ & \footnotesize 15.5\% & \footnotesize 17.6\% & \footnotesize 0.8\% & \footnotesize 0.8\% \\ \cline{2-6}
                  & \footnotesize $K = 400$, $\sigma = 4$ & \footnotesize 23.8\% & \footnotesize 23.4\% & \footnotesize 1.3\% & \footnotesize 2.0\% \\ \hline \hline																		 
									
                  & \footnotesize $K = 100$, $\sigma = 1$ & \footnotesize 0.209 & \footnotesize 0.218 & \footnotesize 0.0225 & \footnotesize 0.0415 \\ \cline{2-6}
\footnotesize AGE & \footnotesize $K = 100$, $\sigma = 4$ & \footnotesize 0.223 & \footnotesize 0.239 & \footnotesize 0.0624 & \footnotesize 0.0606 \\ \cline{2-6}
                  & \footnotesize $K = 400$, $\sigma = 1$ & \footnotesize 0.338 & \footnotesize 0.339 & \footnotesize 0.0592 & \footnotesize 0.0718 \\ \cline{2-6}
                  & \footnotesize $K = 400$, $\sigma = 4$ & \footnotesize 0.345 & \footnotesize 0.348 & \footnotesize 0.0819 & \footnotesize 0.0802\\ \hline									 
									
\hline
\end{tabular}
\end{table}

\bibliographystyle{IEEEtran}
\bibliography{bibfile}

\end{document}